# LMHLD: A Large-scale Multi-source High-resolution Landslide Dataset for Landslide Detection based on Deep Learning

Guanting Liu [a], Yi Wang [a, b, c, d, e, *], Xi Chen [b], Baoyu Du [b], Penglei Li [b], Yuan Wu [a], Zhice Fang [e]

[a] School of Future Technology, China University of Geosciences, Wuhan 430074, China

[b] School of Geophysics and Geomatics, China University of Geosciences, Wuhan 430074, China

[c] Key Laboratory of Ocean Space Resource Management Technology, Ministry of Natural Resources, Hangzhou 310012, China

[d] State Key Laboratory of Resources and Environmental Information System, Beijing 100101, China

[e] Aerospace Information Research Institute, Henan Academy of Science, Zhengzhou 450046, China

***ABSTRACT*--** Landslides are among the most common natural disasters globally, posing significant threats to human society. Deep learning (DL) has proven to be an effective method for rapidly generating landslide inventories in large-scale disaster areas. However, DL models rely heavily on high-quality labeled landslide data for strong feature extraction capabilities. And landslide detection using DL urgently needs a benchmark dataset to evaluate the generalization ability of the latest models. To solve the above problems, we construct a Large-scale Multi-source High-resolution Landslide Dataset (LMHLD) for Landslide Detection based on DL. LMHLD collects remote sensing images from five different satellite sensors across seven study areas worldwide: Wenchuan, China (2008); Rio de Janeiro, Brazil (2011); Gorkha, Nepal (2015); Jiuzhaigou, China (2015); Taiwan, China (2018); Hokkaido, Japan (2018); Emilia-Romagna, Italy (2023). The dataset includes a total of 25,365 patches, with different patch sizes to accommodate different landslide scales. Additionally, a training module, $LMHLD_{part}$, is designed to accommodate landslide detection tasks at varying scales and to alleviate the issue of catastrophic forgetting in multi-task learning. Furthermore, the models trained by LMHLD is applied in other datasets to highlight the robustness of LMHLD. Five dataset quality evaluation experiments designed by using seven DL models from the U-Net family demonstrate that LMHLD has the potential to become a benchmark dataset for landslide detection. LMHLD is open access and can be accessed through the link: https://doi.org/10.5281/zenodo.11424988. This dataset provides a strong foundation for DL models, accelerates the development of DL in landslide detection, and serves as a valuable resource for landslide prevention and mitigation efforts.

## 1 Introduction

Landslides are common hazards in mountainous regions worldwide, often causing extensive damage, economic losses, and casualties (Dai et al., 2002; Fang et al., 2024). Factors such as seismic activity, extreme weather, and rapid urbanization, which contribute to slope instability, are significant drivers of landslides (Guzzetti et al., 1999). Over the past decade, increased geological activity (e.g., volcanic eruptions, earthquakes) and more frequent extreme weather events (e.g., heavy rainfall, hurricanes) have resulted in frequent landslide occurrences globally, posing significant challenges and impacts on human life and production (Doglioni et al., 2020; Muñoz-Torrero Manchado et al., 2021; Saroli et al., 2021; Wang et al., 2023). Therefore, there is an urgent need to rapidly obtain detailed landslide inventories to mitigate the various hazards caused by landslides.

---

[*] Corresponding author.
*E-mail address:* cug.yi.wang@gmail.com (Y. Wang).

Quick monitoring and recording of landslides can help researchers better analyze the causes and impacts of landslides (Guzzetti et al., 2012). Landslide inventories provide information such as occurrence time, location, extent, and magnitude, which can be used to infer the spatial and frequency distribution of landslide events, understand the characteristics of different types of landslides, anticipate future trends, assess potential risks, and formulate effective countermeasures (Luetzenburg et al., 2022; Malamud et al., 2004). Traditional landslide mapping methods include field surveys in geomorphology and visual interpretation of aerial and satellite imagery. However, the former often faces challenges such as difficulty in detecting aging landslides and unclear landslide boundaries, while the latter requires expertise and uniform interpretation standards (Guzzetti et al., 2012).

The rapid development of DL in the field of computer vision has opened up new directions for landslide research. Many researchers have applied DL to landslide inventory mapping and landslide detection using remote sensing images (Azarafza et al., 2021; Bera et al., 2021; Lei et al., 2019). However, the quality of DL models depends on high-quality labeled data (Meena et al., 2023). There is currently a lack of DL datasets applicable to landslide detection or rapid mapping as mentioned in the above and more papers. Many researchers have made efforts to address this critical issue. For example, Ji *et al.* (Ji et al., 2020) interpreted 770 landslides in Bijie, China, and generated a landslide dataset with a spatial resolution of 0.8 m. Luetzenburg *et al.* (Luetzenburg et al., 2022) manually mapped 3,202 landslides in Denmark and made the data publicly available. However, many similar datasets both encompass only one study area, with landslide triggering conditions limited to either rainfall-induced or earthquake-induced. Moreover, the training samples include only vegetation as background besides landslides. This results in incomplete landslide features in the images, and DL models trained on these datasets exhibit poor detection capability for landslides in unknown areas and lack generalization ability. Some recent public landslide datasets have aimed to address the aforementioned issues, including Landslide4Sense (Ghorbanzadeh et al., 2022), HR-GLDD (Meena et al., 2023), and CAS (Xu et al., 2024), with the aim of proposing benchmark datasets for landslide detection to promote the rapid development of DL models.

Undoubtedly, the mentioned landslide datasets have facilitated the applicability of DL models in the field of landslide detection (Asadi et al., 2024; Bravo-López et al., 2022; Chen et al., 2023b; Han et al., 2023). However, to construct a benchmark dataset suitable for landslide detection in DL, it is necessary to address the following limitations: (1) Existing datasets include various landslide-triggering factors and geographical spatial ranges, but the use of satellite sensors with a single spatial resolution limits model robustness and generalization. Due to the significant variability in landslide scales, remote sensing images with a single spatial resolution cannot simultaneously provide both fine-grained and global features, especially using low-resolution satellite sensor cause loss of detail during upsampling or downsampling. Therefore, a dataset with high-resolution multi-source sensor is necessary. (2) For existing landslide datasets, fixed patch sizes (e.g., 128, 512) are typically chosen to segment remote sensing images. However, studies in other fields have been demonstrated that both patch size and spatial resolution significantly influence the performance of DL models, and selecting patch sizes that are tailored to the specific task can effectively improve target detection accuracy (Li et al., 2019; Roy et al., 2019). (3) Compared to photographs, remote sensing images often feature more ground objects, making them more complex to interpret and requiring specialized geological knowledge for landslide annotation. While manual labeling is more time-consuming and labor-intensive than semi-automatic identification methods, it can produce more meticulous and accurate annotations. Therefore, the construction of the landslide benchmark dataset needs to include manual annotation, inspection and correction, and cannot rely entirely on automated processes. (4) Landslide types are diverse, and different types show significant differences in how they are represented in images. As such, a benchmark dataset for experimental research must be scalable, allowing researchers to easily update it by adding new landslide events as they occur. (5) A universal benchmark dataset must undergo thorough experimental validation before being published and put into use (Deng et al., 2009; Hong et al., 2023; Yao et al., 2023).

Based on the above requirements, we construct the Large-scale Multi-source High-resolution Landslide Dataset (LMHLD) to provide a universal benchmark dataset for Landslide Detection based on DL. The main contributions of our research are summarized as follows:

(1) LMHLD is constructed using high-resolution remote sensing images from five distinct commercial satellite sensors, encompassing seven representative landslide study areas worldwide. LMHLD provides data on various types of landslides under different weather or lighting conditions, which can help DL models capture the surface characteristics and changes of different types of landslides and contribute to a comprehensive analysis of the features of landslides in different geographical environments.

(2) An optimal patch sizes selection method is proposed for LMHLD. This selection method mitigates information loss caused by suboptimal patch sizes, thereby enhancing model performance and enabling precise adaptation to landslide features across varying spatial scales.

(3) $LMHLD_{part}$ is proposed to support the seamless integration of multiple sub-datasets within LMHLD. It allows new landslide datasets add into existing models for continuing training, while alleviates the catastrophic forgetting issue commonly encountered in multi-task learning.

(4) Five dataset evaluation experiments are conducted comprehensively to assess the reasonableness, scalability, and overall superiority of LMHLD. The results substantiate the potential of LMHLD as a benchmark dataset, providing reliable inventories for landslide detection.

## 2 Large-scale Multi-source High-resolution Landslide Dataset

### 2.1 Study areas and Satellite sensors

The selection of study areas should consider various influencing factors, including the triggering conditions of landslide events, topography, land cover, and the accessibility of acquiring remote sensing imagery. Through extensive research and analysis of landslide events across multiple areas, the seven study areas and five different types of satellite sensors are selected. For each study area, we compiled multi-scene images sourced from the RapidEye, PlanetScope, IKONOS, Gaofen-2, and Sentinel-2 satellites. The details are shown in Table 1.

The images from each study area were acquired at the most recent available time after the landslides, ensuring the timeliness of the data. The dataset contains approximately 20,000 landslides, with the number of landslides varying across regions based on the differences in landslide distribution and frequency. The pixel coverage of the landslides ranges from as small as 1 pixel to as large as 325,589 pixels, encompassing various types from small-scale to large-scale landslides.

This diverse landslide dataset provides rich training samples for landslide detection and analysis. Additionally, the global distribution of seven study areas is shown in Fig. 1, where the global susceptibility base map was drawn by Tang et al. base on the automated machine (Tang et al., 2023).

Table 1    Study areas in the dataset.

| Study Area | Satellite Sensor | Spatial Resolution | Date | Number of Landslides | Minimum/Maximum Area of Landslide (pixels) |
|---|---|---|---|---|---|
| Rio de Janeiro, Brazil | RapidEye | 5m | 2011.08 | 716 | 7-3124 |
| Hokkaido, Japan | PlanetScope | 3m | 2018.11 | 5666 | 8-22510 |
| Emilia-Romagna, Italy | Sentinel-2 | 10m | 2023.08 | 5847 | 1-4735 |
| Gorkha, Nepal | Sentinel-2 | 10m | 2015.12 | 1760 | 8-6491 |
| Jiuzhaigou, China | Gaofen-2 | 1m | 2015.08 | 3443 | 10-97776 |
| Wenchuan, China | IKONOS | 0.8m | 2008.05 | 4260 | 7-681850 |
| Taiwan, China | PlanetScope | 3m | 2018.01 | 10604 | 32-325589 |

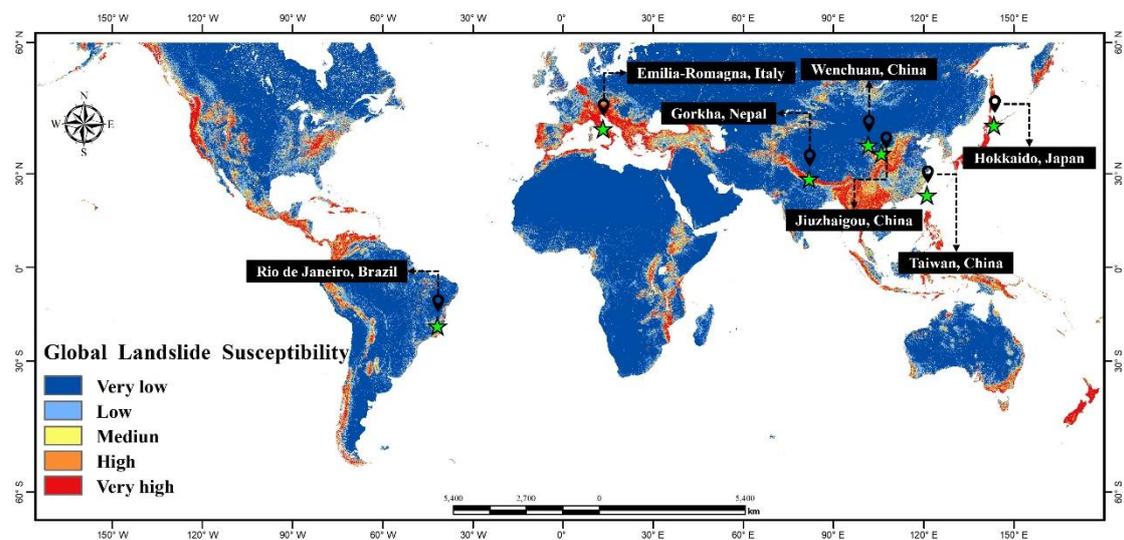

Fig. 1    Location of the study area in the global landslide susceptibility map.

**a) Rio de Janeiro, Brazil**

Rio de Janeiro, located along Brazil's southeastern coast, experiences frequent landslides due to its dense population settlement in steep terrain, exacerbated by rapid urbanization (Ehrlich et al., 2021). Heavy summer rainfall and deforestation further elevate landslide risks (Neto et al., 1999).

**b) Hokkaido, Japan**

Hokkaido, Japan's second-largest island, is prone to landslides due to its seismic activity and geological composition, with volcanic debris contributing to periodic volcanic eruptions (Takashima et al., 2004). The 2018 earthquake triggered numerous landslides, predominantly small to medium-sized, affecting eastern and central Hokkaido (Wang et al., 2019; Zhang et al., 2019).

**c) Emilia-Romagna, Italy**

Emilia-Romagna, central Italy's link between northern and central regions, faces landslide risks due to its diverse topography and climate. Heavy rainfall in May 2023 led to widespread flooding in plains and numerous landslides in hilly areas (Ferrario and Livio, 2023).

**d) Gorkha, Nepal**

Gorkha, Nepal's central region, is susceptible to landslides due to its mountainous terrain and seismic activity. The 2015 earthquake triggered thousands of landslides, affecting a significant area (Martha et al., 2017; Roback et al., 2018).

**e) Jiuzhaigou, China**

Jiuzhaigou, located in northern Sichuan Province, China, is characterized by high-altitude karst landforms. The region experienced a 7.0 magnitude earthquake on August 8, 2017, triggering over 1780 landslides of various types, including shallow slides and rockfalls (Tian et al., 2019).

### f) Wenchuan, China

Wenchuan, situated in northern Sichuan Province, China, is prone to seismic activity due to its active tectonic zones. The 2008 Wenchuan earthquake, with a magnitude of Mw8.0, resulted in approximately 15,000 landslides and caused significant casualties and economic losses (Yin et al., 2009).

### g) Taiwan, China

Taiwan, an island off the southeastern coast of China, faces frequent natural disasters such as earthquakes and typhoons. Its mountainous terrain, combined with heavy rainfall, contributes to landslide susceptibility (Huang et al., 2017).

## 2.2 Landslide inventory production

Creating a landslide inventory is crucial for assessing landslide susceptibility and hazard (Kirschbaum et al., 2015), and a good landslide inventory provides valuable data support for landslide detection based on supervised DL algorithms. Current methods for creating high-precision landslide inventories, such as field surveys (Brunsden, 1985), manual interpretation of images (Luetzenburg et al., 2022), or data retrieval from technical reports or newspapers (Franceschini et al., 2022) mostly rely on subjective judgments and are time-consuming. The lack of geological expertise often leads to misjudgments, resulting in inaccurate inventories. This undermines the reliability of detection models and limits their real-world applicability. To avoid these problems, we have implemented the following measures method to ensure the accuracy and reliability of landslide inventory annotations.

Firstly, existing landslide inventory data from different areas, such as the multi-temporal landslide inventory for the entire province of Taiwan released by the Taiwan Council of Agriculture since 2004[1], the landslide inventory conducted by the Institute for Environmental Protection and Research (ISPRA) of Italy (IFFI project) - *Geological Survey of Italy* (Trigila and Iadanza, 2007), *etc*, is referenced during the annotation process. Comparing landslide inventory data with images from a similar timeframe can provide accurate references, as vegetation recovery after landslides takes time.

Landslide-associated phenomena can serve as valuable auxiliary information for image interpretation. The landslide area appears brighter than the surrounding non-landslide areas in images, as the surface cover is disrupted, exposing fresh soil and rock beneath after a landslide disaster, as shown in Fig.2 (a). Furthermore, the associated geological structures and subsequent phenomena, such as changes in vegetation or surface features, can also serve as important indicators for landslide annotations (Varnes, 1958). The vegetation on the surface after a landslide is blocky and fragmented, as shown in Fig.2 (d). Landslide events lead to extensive loss of surface vegetation, and the near-infrared bands in remote sensing images, which capture the 'red edge' phenomenon of vegetation (Horler et al., 1983), are particularly useful for landslide labeling. When the near-infrared band is added for color display, non-landslide areas with vegetation cover appear red. This helps in differentiating landslides from the background, as shown in Fig. 2(e). The debris flow formed by the flow of surface cover along the slope after a landslide occurs, as shown in Fig.2 (c). And the sliding mass of a landslide typically forms accumulation zones downslope, as shown in Fig.2 (d). All of these are the image features we reference during the landslide annotation process.

---

[1] https://data.gov.tw/

Finally, landslide occurrences not only require triggering conditions such as earthquakes and rainfall but also depend on slope and elevation differences. Therefore, in the manual annotation process, we actively reference the slope and elevation data to minimize misclassifications.

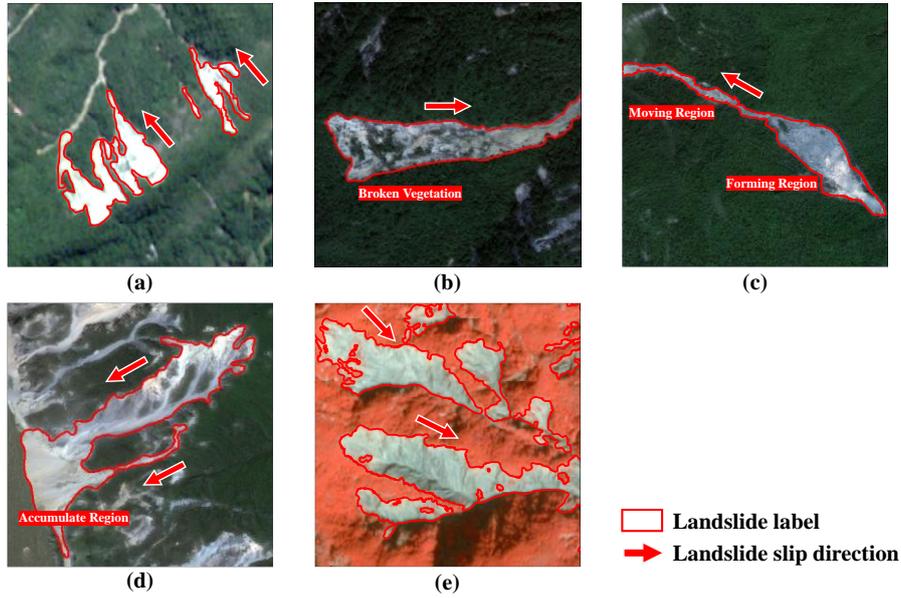

Fig. 2    Image features in the process of landslides annotation.

## 2.3 Dataset construction

Selecting patch sizes tailored to the specific task can effectively improve target detection accuracy (Li et al., 2019; Roy et al., 2019). Meena *et al.*(Meena et al., 2022) also pointed out that using different patch sizes for cropping remote sensing images for landslide detection yields different performance across models. Excessively large image sizes can significantly increase the computational complexity, which in return reduces training efficiency (Ball et al., 2017).

Consideration above, we propose a new patch size selection method for landslide detection dataset. The method crops images of different patch sizes based on the actual distribution of landslide scales in different study areas and the spatial resolution of the image to preserve target and context information for different scales of landslides. And the patch sizes selected for different study areas are shown in Table 2 (the selection of optimal patch sizes is described in **Section 3**).

Therefore, cropping into patches has become an important preprocessing step for remote sensing images (Cheng et al., 2020). The dataset in this study consists of remote sensing images from multiple study areas globally, with significant variations in landslide scales even within the same study area, as shown in Table 1. To preserve context information for different scales of landslides, we crop images of different patch sizes based on the actual distribution of landslide scales in different study areas and the spatial resolution of the image. Meena *et al.*(Meena et al., 2022) pointed out that using different patch sizes to crop remote sensing images for landslide detection shows different performance in the results of models. In this study, the patch sizes selected for different study areas are shown in Table 2 (the selection of optimal patch sizes is described in **Section 3**).

The entire image is cropped by sliding the window with the patch size as the step, ensuring no overlap between adjacent patches. The sliding window segmentation method is inherently random, which may result in patches with very few landslide pixels, leading to an imbalance between positive and negative samples. Additionally, when the sliding window is applied at the image boundary, the resulting patches may contain large areas of 'no data', potentially causing anomalies in the dataset during subsequent experiments. To address these issues, we further eliminate patches that do not contain landslide pixels, aiming to balance the positive and negative samples. Specifically, after segmentation, we apply the following filtering rules: (1) remove patches where landslide pixels account for less than 10% of the total patch area (Ghorbanzadeh et al., 2022; Xu et al., 2024); (2) remove patches that contain image boundaries. Through these processes, we ensure that the selected patches are more representative. Each selected patch includes four spectral bands (blue, green, red, and near-infrared) to ensure comprehensive information is retained for effective model training.

Table 2    Patch size and number of patches in different study areas.

| Study Area | Patch Size | Number of Patches | Proportion of Landslide Pixels | | |
|---|---|---|---|---|---|
| | | | Train set | Validation set | Test set |
| Rio de Janeiro, Brazil | 64 | 683 | 0.1744 | 0.1693 | 0.1768 |
| Hokkaido, Japan | 128 | 1915 | 0.2494 | 0.2534 | 0.2519 |
| Emilia-Romagna, Italy | 32 | 5929 | 0.2004 | 0.1953 | 0.1946 |
| Gorkha, Nepal | 32 | 3972 | 0.2597 | 0.2578 | 0.2508 |
| Jiuzhaigou, China | 128 | 2734 | 0.2388 | 0.2436 | 0.2475 |
| Wenchuan, China | 224 | 2603 | 0.3243 | 0.3227 | 0.3239 |
| Taiwan, China | 224 | 7529 | 0.2106 | 0.2130 | 0.2080 |
| Total | - | 25365 | - | - | - |
| Comparison dataset | 128 | 28185 | 0.2968 | 0.2895 | 0.2924 |

Additionally, we divide 70% of all sub-datasets into the training dataset, 20% into the validation dataset, and 10% into the testing dataset. And the proportion of landslide pixels in each of the training, validation, and testing datasets, as shown in Table 2. The difference in the proportion of landslide pixels among the three datasets does not exceed 1%, which can mitigate the problem of class imbalance and help the model generalize better. And we extra constructed a comparison dataset, using a patch size of 128 for comparison to accommodate a larger capacity DL model and verify the importance of patch size selection.

We compare LMHLD with five public landslide datasets to comprehensively assess its uniqueness and superiority. As shown in Table 3, LMHLD demonstrates significant advantages in labeled data volume, patch size and sensors diversity. These strengths are further validated by the experimental results presented in Section 3 and 4.

Table 3 Comparison of LMHLD with five public landslide datasets.

| Dataset | Patch Size | Satellite Sensor | Resolution | Number of Labeled Patches |
|---|---|---|---|---|
| Bijie landslide | - | TripleSat | 0.8m | 2773 |
| Landslide4sense | 128 | Sentinel-2 | 10m | 3799 |
| HR-GLDD | 128 | PlanetScope | 3m | 1756 |
| GVLM | 256 | SPOT-6 | 1.5m | 630 |
| CAS Landslide (Only SAT) | 512 | PlanetScope, SuperView-1, GF-1, WorldView2/3, Sentinel-2, Landsat | 0.5m-15m | 7422 |
| **LMHLD** | **32, 64, 128, 224** | **RapidEye, PlanetScope, GF-2, IKONOS, Sentinel-2** | **0.8m-10m** | **25365** |

## 3 Methodology and Experiments

### 3.1 Model architectures

As a pivotal benchmark model in the field of image semantic segmentation within DL technology, U-Net (Ronneberger et al., 2015) and its variants have been widely applied in landslide detection, achieving outstanding performance (Bhuyan et al., 2023; Chen et al., 2023a; Dong et al., 2022; Meena et al., 2022; Niu et al., 2022; Qi et al., 2020). U-Net, with encoder-decoder structure, possesses strong feature extraction, resolution restoration, and cross-scale information fusion capabilities. It is inherently compatible with CNN architectures and can also be integrated with advanced structures such as Transformer and Mamba. Additionally, its decoder employs progressive upsampling and skip connections, aligning with other DL semantic segmentation methods such as FCN, DeepLabV3+, and HRNet (Chen et al., 2018; Shelhamer et al., 2016; Wang et al., 2020).

In this study, we selected U-Net, Res U-Net (Xiao et al., 2018), Dense U-Net (Guan et al., 2020), Attention U-Net (Oktay et al., 2018), Trans U-Net (Chen et al., 2021), U-Net++ (Zhou et al., 2018), and U-Net+++ (Huang et al., 2020) as the training networks. These seven U-Net variants represent key innovation trends in DL, spanning from CNN-based structural optimizations to the incorporation of Transformer architectures, as well as advancements such as residual connections, skip connections, and attention mechanisms. The evolution of U-Net and its variants illustrates the broader trajectory of deep learning in semantic segmentation, which primarily focuses on developing more advanced models with deeper feature extraction, stronger attention mechanisms, enhanced feature fusion, and improved global information modeling.

Therefore, we argue that U-Net and its variants encapsulate the major innovative directions of DL in landslide detection, and we adopt the U-Net family as the feature extraction model to conduct experiments on the LMHLD dataset to evaluate the scientific validity and appropriateness of LMHLD as a benchmark DL dataset for landslide detection in remote sensing imagery.

### 3.2 Experimental design

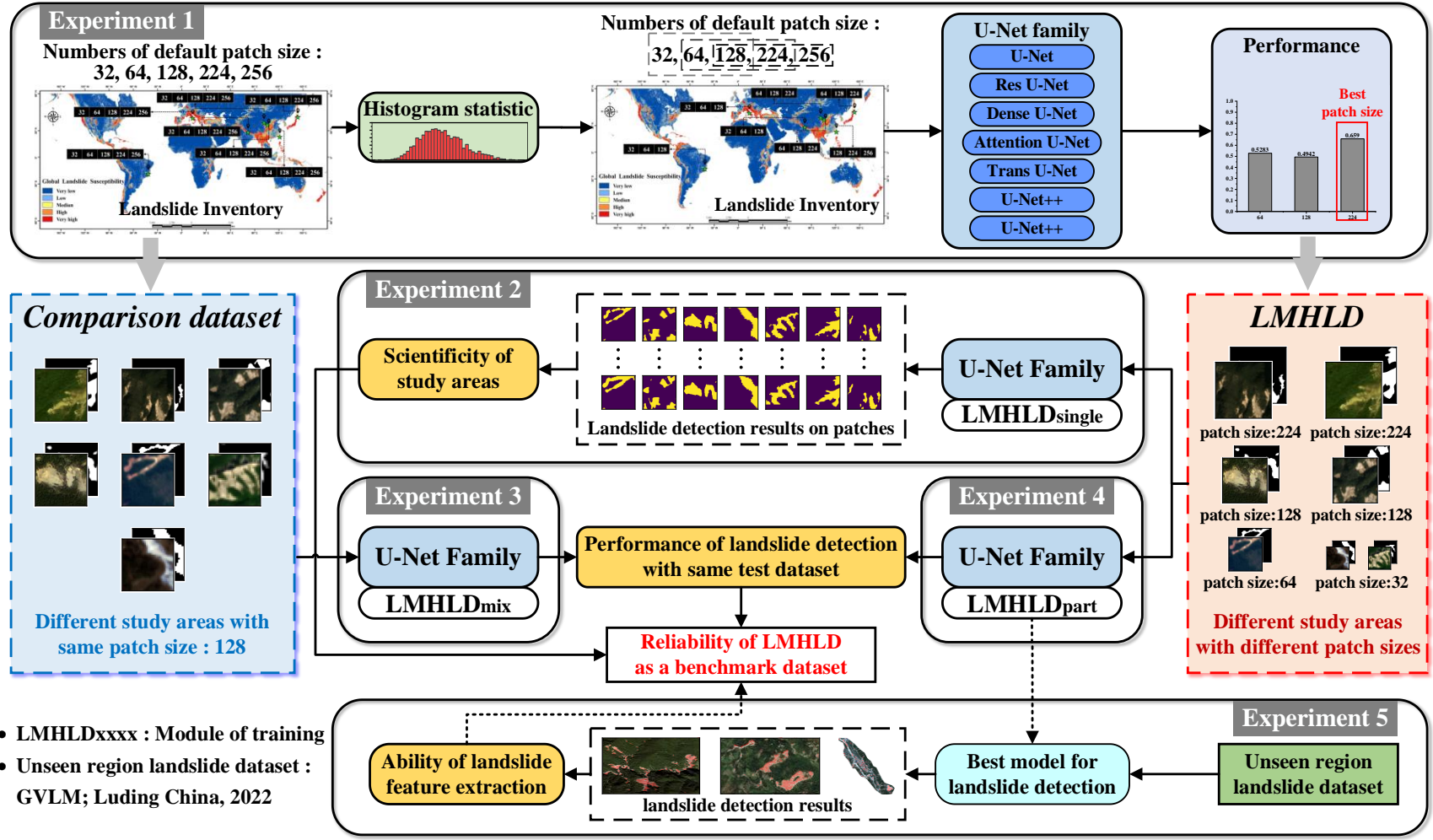

**Fig. 3** Flowchart of the whole experimental scheme.

As mentioned in **Section 1**, a landslide benchmark dataset must undergo rigorous experimental verification to ensure its reliability and applicability in real-world scenarios. The design of the experimental scheme primarily revolves around evaluating "the scientific validity and reliability of LMHLD as a benchmark dataset for landslide detection". Experiments are conducted on LMHLD, training the U-Net Family models, which include U-Net, Attention U-Net, Res U-Net, Dense U-Net, U-Net++, U-Net+++, and Trans U-Net. We design three experimental modules ($LMHLD_{single}$, $LMHLD_{mix}$ and $LMHLD_{part}$) to optimize the model training process across different validation scenarios. The flowchart of the whole experimental scheme is shown in Fig. 3. First, the optimal patch size for image cropping is selected based on the characteristics of the study area. Next, the reliability of LMHLD is further validated by evaluating the performance of different networks on the selected areas. This helps to assess whether these areas are representative and whether the model can be effectively trained and tested. Finally, the model trained by LMHLD is applied to detect landslides in unseen areas, which helps verify whether LMHLD can provide the necessary generalization and robustness for DL models.

(1) *Experiment* 1: This part of the experiment focuses primarily on selecting the optimal patch size for different study areas. The default patch sizes are 32, 64, 128, 224, 256. We performed histogram statistics of landslide scale (in terms of pixel counts) using the landslide inventories from seven study areas. From this, we calculated the first quartile (Q1), the second quartile (Q2), also called the median and the third quartile (Q3) (Hyndman and Fan, 1996), to assess the distribution and dispersion degree of landslide scale in each study area. Subsequently, the default patch size range is narrowed to an interval of three values, and seven DL models are used for training. The performance of landslide detection across the entire study area serves as the criterion for selecting the optimal patch size.

(2) *Experiment* 2: Scientific validation of seven sub-region dataset. After determining the optimal patch size, we constructed corresponding sub-datasets. Each sub-dataset was divided into training, validation, and test sets, which were then input into seven U-Net family models for training and testing. This process is called $LMHLD_{single}$. The model's performance on the test set can evaluate whether the landslide dataset LMHLD, constructed based on these seven regions, is representative and whether it provides effective training and testing data for the models. This, in turn, allows for further assessment of the scientific validity and reliability of the dataset LMHLD.

(3) *Experiment* 3: Scientific validation of comparation dataset. The main difference between comparation dataset and LMHTD lies in the use of a consistent 128 patch size, whereas LMHTD utilizes varying patch sizes. To evaluate comparation dataset's performance, we apply seven DL models from the U-Net family for training and testing. This process is called $LMHLD_{mix}$, as shown in Fig. 4. The model's performance on the test set can assess the impact of using a specific patch size on landslide detection performance. And the results will be compared with those from Experiment 4 to explore how patch size selection influences landslide detection performance across different study areas and scales.

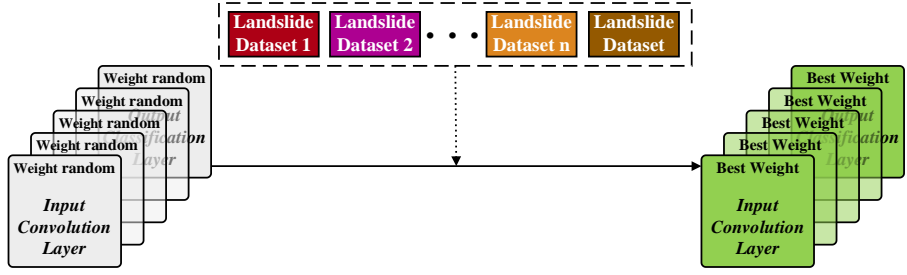

**Fig. 4    The experimental strategy using LMHLD$_{mix}$.**

(4) *Experiment* 4: Scientific validation of serialized training strategy. To evaluate the performance of serialized training strategy, we apply seven DL models from the U-Net family for training and testing. This process is called LMHLD$_{part}$, as shown in Fig. 5.

In landslide detection method based on DL, usually requires retraining the entire model when applied to a new dataset, causing time-consuming especially dealing with large datasets. To address this challenge, we propose a serialized training strategy that enables batch training on heterogeneous datasets (with varying resolutions and patch sizes) while effectively mitigating catastrophic forgetting (Kirkpatrick et al., 2017).

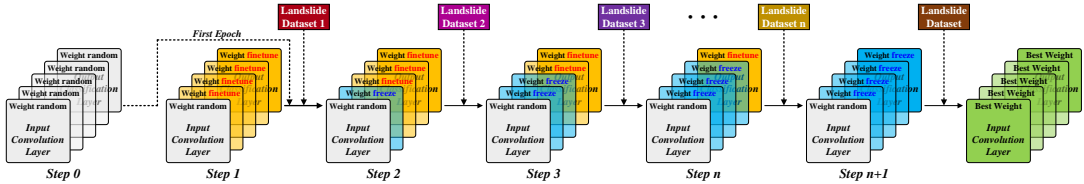

**Fig. 5    The experimental strategy using LMHLD$_{part}$.**

During this process, we employ a serialized training strategy, where multiple landslide datasets (Landslide Dataset 1, Dataset 2, ..., Dataset n) are gradually introduced for sequential training. In each training phase, a new dataset is added to further optimize the model, while selectively freezing and updating network parameters during the training process. The specific process is as follows:

In the first epoch, all network parameters are randomly initialized (marked as 'weight random' in Fig.5), ensuring that the model starts learning landslide features from scratch. Starting from the second epoch, the serialized training strategy is applied, gradually adding sub-datasets. Each epoch a new dataset is introduced, the model continues training from the parameters of the previous epoch and uses them as the initial weights for the new dataset. This transfer strategy leverages previously learned landslide knowledge, accelerating convergence and improving detection performance.

During training, due to the hierarchical structure of the network, different layers of the model perform different parameter optimization strategies. First, the weights of the input convolutional layers are randomly initialized to adapt to the shape of the new dataset. Shallow layers that capture low-level features (such as edges and shapes) are typically universal. Therefore, as training progresses, the shallow layers are gradually frozen to prevent overfitting and retain the previously learned landslide features. These layers are marked as 'weight frozen' in Fig.5. Deeper features, which are often incomplete, continue to train (marked as 'weight finetune') and update their parameters to adapt to different data characteristics, enhancing the model's ability to generalize across various types of landslides. After the entire training process is completed, a pre-trained model is obtained, which can be used for downstream landslide detection tasks in fixed regions.

To reduce the experimental complexity and time costs, the order of dataset addition is fixed as Rio de Janeiro, Brazil; Gorkha, Nepal; Emilia-Romagna, Italy in the first three times of adding sub-datasets, while the final addition comes from Wenchuan, China. In addition, three other areas

(Hokkaido, Japan / Taiwan, China / Jiuzhaigou, China) are selected for training, with each selection being unique. The six different sequences of data addition and training are shown in Fig. 6. The experimental conclusions can also be extended to the case where the seven sub-datasets are added in different orders.

The model's performance on the test set can assess the performance of LMHLD$_{part}$ on landslide detection performance. And the results compared with Experiment 4 can further valid scientific of LMHLD with different patch size.

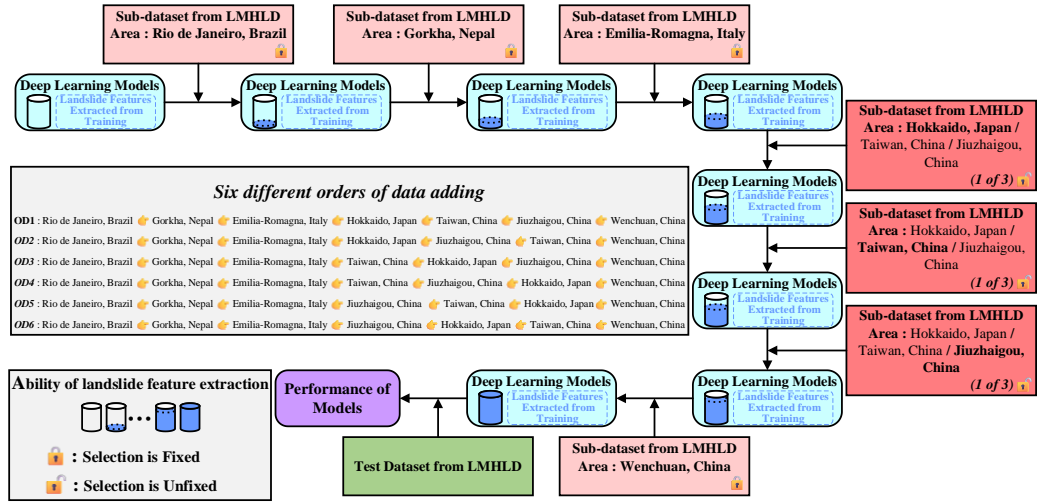

Fig. 6　　Six different orders of data addition during LMHLD$_{part}$.

(5) *Experiment* 5: Generalization test. To evaluate the generalization and detection performance of the model trained with LMHLD on unseen landslides, we use two additional datasets: the public landslide dataset GVLM (Zhang et al., 2023) and a recent landslide image triggered by an earthquake in Luding in 2022, collected from Planet data (An et al., 2023). The images used for testing are shown in Fig.7 and 8. This experiment can further determine LMHLD, as a benchmark landslide detection dataset, can provide reliable and robust general landslide features for the latest DL models.

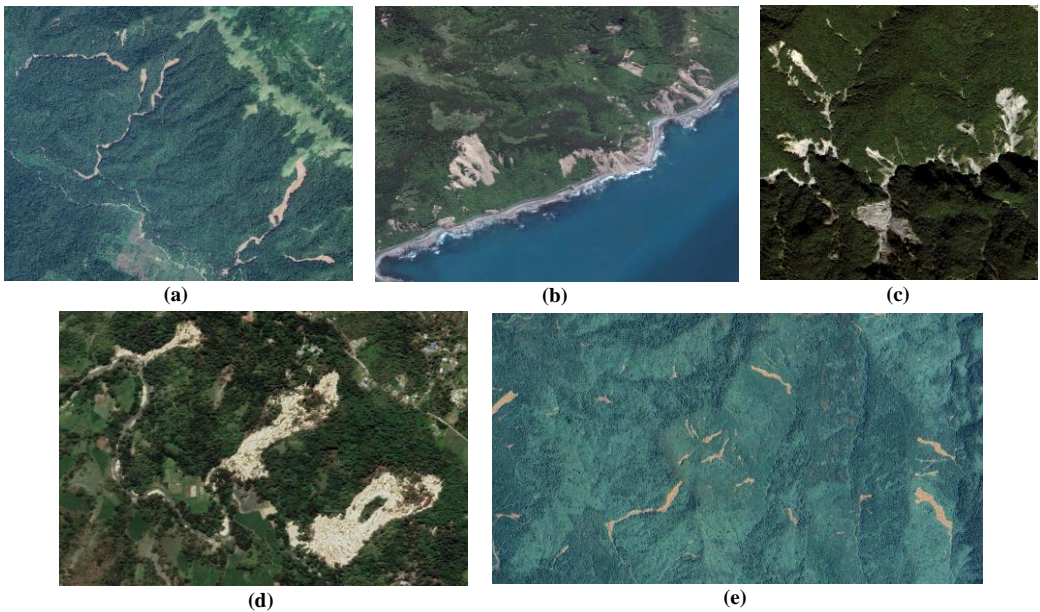

Fig. 7　　The test data selected from GVLM. (a) Kodagu_India. (b) Kaikoura_New Zealand. (c) Taitung_China. (d) Kupang_Indonesia. (e) ALuoi_Vietnam.

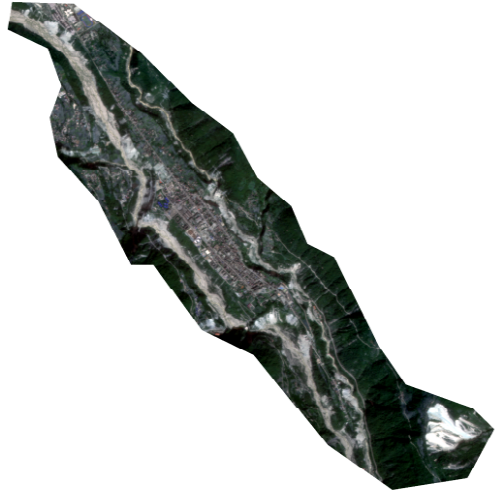

**Fig. 8** The remote sensing image of Luding, Sichuan Province, China, which is composed of planet on October 21th, 2022.

**3.3 Experimental settings**

In this study, all experiments were conducted on a cluster equipped with four NVIDIA RTX 3090 GPUs. We employed the P yTorch framework to construct seven DL models. To facilitate efficient training of these models, we utilized the "Xavier" method (Glorot and Bengio, 2010) to initialize the convolutional layers, linear layers, and normalization layers within the networks. All the models were optimized using stochastic gradient descent (SGD) optimizer, with an initial learning rate set to $1.00×10^{-2}$ and an exponential learning rate decay with a gamma value of 0.99. Considering hardware performance and time efficiency, we conducted several experiments to evaluate different combinations of batch sizes (8, 16, 32, 64) and training epochs (50, 100, 200, 300). A batch size of 16 and 300 epochs provided the best performance, which was selected as the final training configuration.

**3.4 Evaluation metrics**

To assess the performance of different models on the LMHLD and the superiority of the dataset, we use three commonly used evaluation metrics in landslide detection (Li et al., 2023; Shi et al., 2020): Precision, Recall and F1-score. These metrics are calculated based on the true positives (TP), true negatives (TN), false positives (FP), and false negatives (FN) from the confusion matrix. Precision measures the proportion of correctly predicted positive samples among all predicted positive samples. Recall, or sensitivity, evaluates the proportion of correctly predicted positive samples among all actual positive samples. The F1 score, the harmonic mean of precision and recall, offers a balanced measure of both metrics. The formulas for calculating these three metrics are as follows:

$$Precision = \frac{TP}{TP+FP} \quad (1)$$

$$Recall = \frac{TP}{TP+FN} \quad (2)$$

$$F_1\text{-}score = \frac{Precision*Recall}{Precision+Recall} \quad (3)$$

where *TP* is the instances correctly predicted as positive; *TN* is the instances correctly predicted as negative; *FP* is the instances incorrectly predicted as positive; *FN* is the instances incorrectly predicted as negative.

## 4 Results and Discussion

### 4.1 Selection of optimal patch size

Using China's Wenchuan as a case study, we present the process of selecting optimal patch sizes, which follows the same procedure applied to other study areas. As shown in Fig. 9, in Wenchuan, 75% of the landslides have fewer than 4123 pixels but more than 417 pixels, and 50% of landslides are smaller than 1191 pixels. To ensure the proportion of landslide pixels in the patches segmented by sliding windows is as close as possible to 10%, the ranges covered by the windows should approximately be 4170, 11910, and 41230 pixels, respectively. Based on these observations, we initially determine the candidate patch sizes for Wenchuan initially determine as 64, 128, and 224.

Using these three preselected patch sizes, three sub-datasets are created for Wenchuan. Table 4 displays the performance of the seven DL models trained on these three sub-datasets, tested across the entire Wenchuan area. The results show that the 224-patch size yields the highest F1-score, providing more accurate landslide features and reducing misjudgments and misdetections across multiple models. Therefore, we choose 224 as the optimal patch size for this region.

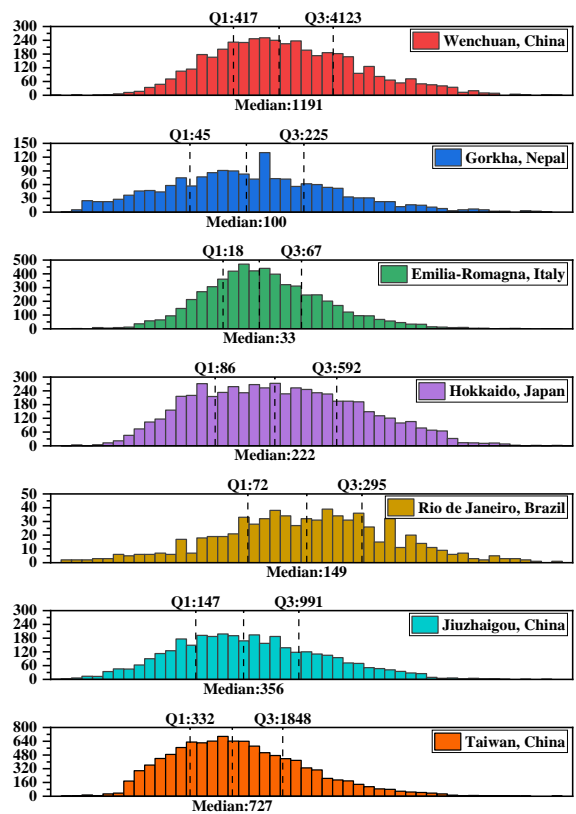

**Fig. 9** The histogram of landslide scale (the numbers of pixels) on the landslide inventory of seven areas. Y-axis: the numbers of landslides of the given scale.

**Table 4** Performance of seven deep learning models with different patch sizes on the Wenchuan sub-dataset.

| Model | Patch Size | Precision | F1-score | Recall |
|---|---|---|---|---|
| U-Net | 64 | 0.3434 | 0.4748 | 0.7693 |
|  | 128 | **0.5305** | 0.6333 | 0.7856 |
|  | 224 | 0.5234 | **0.6486** | **0.8522** |
| Res U-Net | 64 | **0.4459** | **0.5821** | **0.8380** |
|  | 128 | 0.4237 | 0.5607 | 0.8288 |
|  | 224 | 0.3934 | 0.5331 | 0.8264 |
| Dense U-Net | 64 | 0.4577 | 0.6044 | **0.8897** |
|  | 128 | **0.4942** | **0.6171** | 0.8213 |
|  | 224 | 0.4652 | 0.5988 | 0.8401 |
| Attention U-Net | 64 | 0.3967 | 0.5408 | 0.8493 |
|  | 128 | 0.3783 | 0.5398 | **0.9418** |
|  | 224 | **0.5944** | **0.7156** | 0.8990 |
| Trans U-Net | 64 | 0.3823 | 0.5283 | 0.8548 |
|  | 128 | 0.3435 | 0.4942 | **0.8804** |
|  | 224 | **0.5715** | **0.6590** | 0.7782 |
| U-Net++ | 64 | 0.5078 | 0.6483 | 0.8962 |
|  | 128 | 0.3782 | 0.5384 | **0.9343** |
|  | 224 | **0.5933** | **0.6980** | 0.8475 |
| U-Net+++ | 64 | 0.4432 | 0.5519 | 0.7313 |
|  | 128 | 0.5434 | 0.6807 | 0.9106 |
|  | 224 | **0.5679** | **0.6997** | **0.9111** |

**4.2 Scientific validation of seven study areas**

The landslide detection performance of the seven models on the test data from each sub-dataset, based on evaluation metrics such as Precision, Recall, and F1-score, is shown in Table 5. U-Net++ outperforms all other models, achieving the highest F1 score in the study areas of Rio de Janeiro (Brazil), Kathmandu (Nepal), and Wenchuan (China). Trans U-Net, Attention U-Net, and U-Net also perform competitively. Notably, although Res U-Net achieves high precision across four areas, its recall is lower, suggesting the model is more conservative in detecting landslides. Furthermore, apart from U-Net++, the models perform best in Wenchuan, China, achieving the highest F1-score of 0.8802, indicating the high data quality of the Wenchuan area for landslide detection with stable results across models.

The average Precision, F1-score, and Recall of the seven models across the seven study areas for landslide detection tasks are 0.8275, 0.8032, and 0.7852, respectively. The consistently high F1-scores, without significant fluctuations, indicate the reliability and stability of DL models in landslide detection across these study areas. When Precision exceeds Recall, it suggests that the models are better at avoiding false positives, but this may also indicate incomplete detection of boundaries or internal regions in the landslide results. Additionally, a representative image patch from each study area, along with the corresponding ground truth, is shown in Fig. 10 to visually demonstrate the detection results.

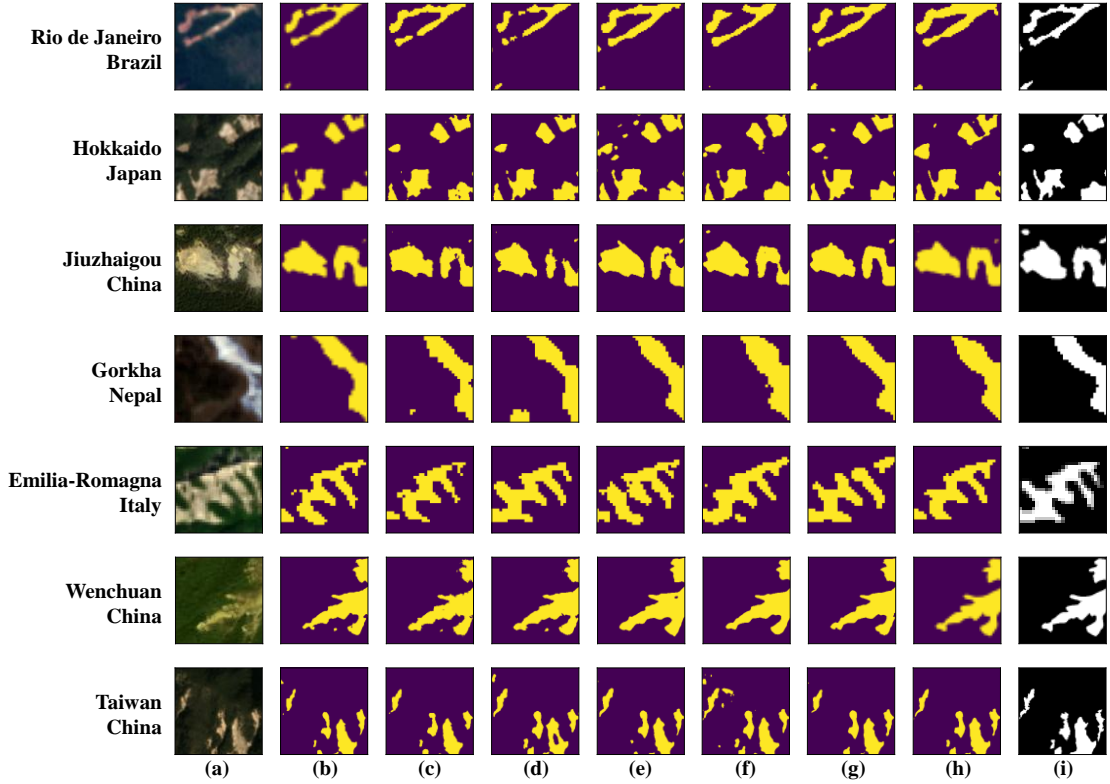

**Fig. 10** Landslide detection results of seven study areas by the training module-LMHLD$_{part}$. (a) Original Image. (b) U-Net. (c) Res U-Net. (d) Dense U-Net. (e) Attention U-Net. (f) Trans U-Net. (g) U-Net++. (h) U-Net+++. (i) Ground Truth.

**Table 5** Performance of seven deep learning models on each sub-dataset of different areas on LMHLD$_{single}$. Values in bold represent the best-performing model for each evaluation metric, while underlined values denote the best-performing study area for each model's evaluation metric.

| Study Area | Evaluation Metric | U-Net | Res U-Net | Dense U-Net | Attention U-Net | Trans U-Net | U-Net++ | U-Net+++ |
|---|---|---|---|---|---|---|---|---|
| Rio de Janeiro Brazil | Precision | 0.8334 | **0.8771** | 0.8643 | 0.7355 | 0.8186 | 0.8072 | 0.7088 |
|  | F1-score | 0.8013 | 0.7704 | 0.7859 | 0.7592 | 0.7683 | **0.8056** | 0.7914 |
|  | Recall | 0.7717 | 0.6869 | 0.7205 | 0.7845 | 0.7238 | 0.8040 | <u>0.8956</u> |
| Hokkaido Japan | Precision | 0.8518 | <u>**0.8940**</u> | 0.8852 | 0.7805 | 0.7960 | 0.8242 | 0.7987 |
|  | F1-score | **0.8134** | 0.7783 | 0.7970 | 0.7943 | 0.8045 | 0.8072 | 0.8027 |
|  | Recall | 0.7783 | 0.6891 | 0.7247 | 0.8087 | **0.8133** | 0.7910 | 0.8068 |
| Emilia-Romagna Italy | Precision | 0.7997 | **0.8613** | 0.8249 | 0.8014 | 0.7951 | 0.8115 | 0.8309 |
|  | F1-score | 0.7928 | 0.7726 | **0.8044** | 0.8008 | 0.7918 | 0.8017 | 0.7812 |
|  | Recall | 0.7860 | 0.7004 | 0.7848 | **0.8001** | 0.7886 | 0.7922 | 0.7372 |
| Gorkha Nepal | Precision | 0.7869 | **0.8474** | 0.8253 | 0.7892 | 0.7740 | 0.7919 | 0.7726 |
|  | F1-score | 0.7914 | 0.6709 | 0.7930 | 0.7907 | 0.7885 | **0.8101** | 0.7695 |
|  | Recall | 0.7958 | 0.5553 | 0.7632 | 0.7923 | 0.8035 | **0.8291** | 0.7663 |
| Jiuzhaigou China | Precision | 0.8100 | 0.8627 | **0.8805** | 0.8191 | 0.8360 | 0.8124 | 0.8156 |
|  | F1-score | 0.8084 | 0.7413 | 0.7443 | **0.8127** | 0.7923 | 0.8028 | 0.8101 |
|  | Recall | **0.8068** | 0.6499 | 0.6446 | 0.8063 | 0.7529 | 0.7934 | 0.8046 |
| Wenchuan China | Precision | 0.8591 | 0.8042 | 0.8460 | <u>0.8721</u> | <u>**0.9195**</u> | <u>0.8865</u> | <u>0.8643</u> |
|  | F1-score | <u>0.8622</u> | <u>0.8223</u> | <u>0.8508</u> | <u>0.8710</u> | <u>0.8335</u> | <u>**0.8802**</u> | <u>0.8667</u> |
|  | Recall | <u>0.8654</u> | <u>0.8411</u> | <u>0.8556</u> | <u>0.8700</u> | 0.7622 | <u>**0.8741**</u> | <u>0.8690</u> |
| Taiwan China | Precision | <u>**0.8735**</u> | 0.8369 | 0.8461 | 0.8502 | 0.7617 | 0.8596 | 0.8455 |
|  | F1-score | 0.8515 | 0.7951 | 0.8148 | **0.8524** | 0.8116 | 0.8512 | 0.8467 |
|  | Recall | 0.8306 | 0.7572 | 0.7858 | 0.8546 | **0.8686** | 0.8430 | 0.8479 |

In conclusion, we demonstrate that the selection of study areas for constructing LMHLD is both rational and scientifically grounded. Moreover, the high-resolution remote sensing images and label data from these areas offer rich and diverse landslide features, enabling DL models to effectively adapt to complex landslide detection tasks.

**4.3 Performance of dataset with the same patch size on LMHLD$_{mix}$**

The performance of the seven DL models on the LMHLD$_{mix}$ for landslide detection is presented in Table 6. Comparing the LMHLD$_{mix}$ results in Table 6 with the results for Japan's Hokkaido and China's Jiuzhaigou (both with a patch size of 128) in Table 5, it can be observed that the DL models perform better in landslide detection when using multi-satellite sensors high-resolution remote sensing images across multiple study areas. This highlights the importance of constructing LMHLD with images from diverse study areas and satellite sensors, each with varying spatial resolutions. By incorporating such diversity in the dataset, the model is better equipped to learn from a wide range of geographical features, environmental conditions, and sensor-specific characteristics. This ensures that the DL models trained on LMHLD are capable of handling real-world variations and complexities in landslide detection, enhancing their robustness and generalization across different areas and sensor types.

Although incorporating different types of images improved the performance of landslide detection in China's Wenchuan and Taiwan (both with a patch size of 224), the enhancement in F1-score was not substantial under conditions where patch size selection was inappropriate. In some cases, only marginal improvements were observed, while for certain models, the evaluation metrics actually declined. Similar patterns were seen during the testing phase, when applying the test datasets from each study area's sub-dataset. These findings further underscore the critical importance of selecting the appropriate patch size for optimal performance.

The visualization results of landslide detection on LMHLD$_{mix}$, across different deep models, are shown in Fig. 11. These visualizations provide an intuitive representation of the model's effectiveness in detecting landslides across various study areas.

**Table 6** Performance of seven deep learning models on LMHLD$_{mix}$. Values in bold indicate the maximum value for each evaluation metric.

| Model | Precision | F1-score | Recall |
| --- | --- | --- | --- |
| U-Net | 0.8587 | 0.8544 | 0.8501 |
| Res U-Net | 0.8693 | 0.8442 | 0.8206 |
| Dense U-Net | 0.8575 | 0.8578 | 0.8581 |
| Attention U-Net | 0.8444 | 0.8552 | 0.8664 |
| Trans U-Net | **0.8732** | **0.8698** | **0.8665** |
| U-Net++ | 0.8608 | 0.8540 | 0.8473 |
| U-Net+++ | 0.8719 | 0.8555 | 0.8441 |

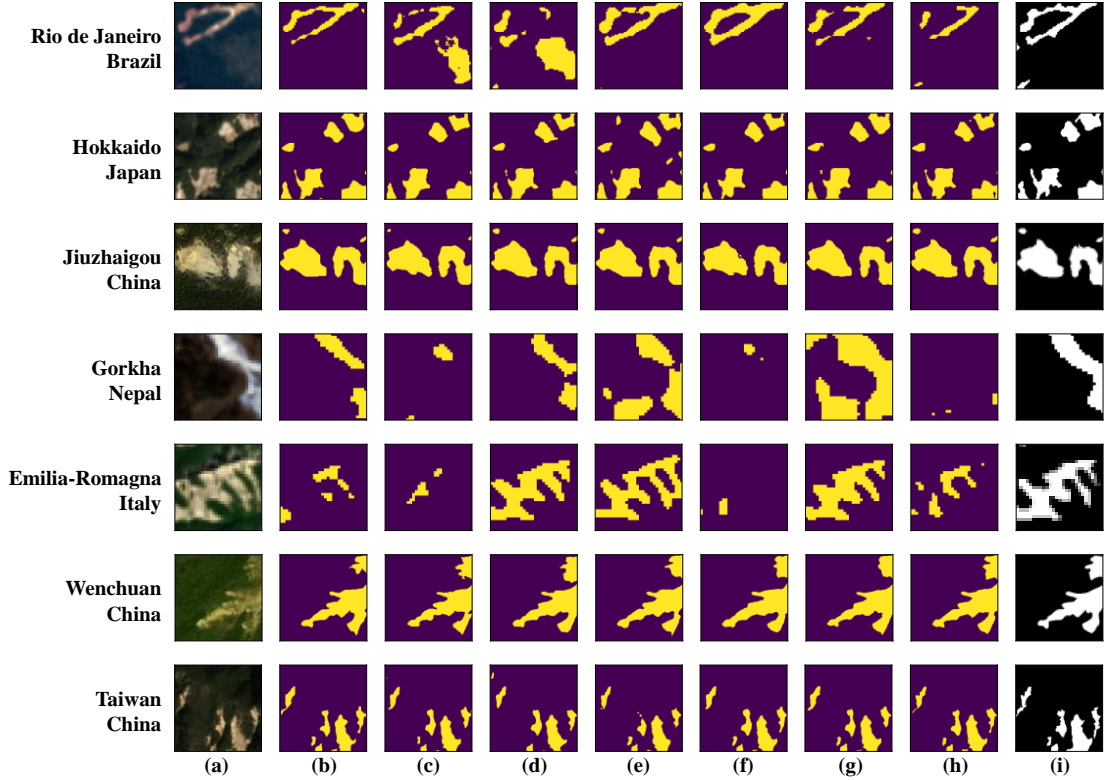

Fig. 11 The landslide detection results of seven study areas by the training module-LMHLD$_{mix}$. (a) Original Image. (b) U-Net. (c) Res U-Net. (d) Dense U-Net. (e) Attention U-Net. (f) Trans U-Net. (g) U-Net++. (h) U-Net+++. (i) Ground Truth.

**4.4 Performance of dataset with different patch sizes on LMHLD$_{part}$**

The precision, F1-score, and recall of the seven DL models on LMHLD$_{part}$, based on different data addition orders, are presented in Figs. 12, 13 and 14. The X-axis and Y-axis represent data with different addition orders. The color gradient, transitioning from white to black with a yellow-red intermediate range, visually illustrates the absolute differences in the evaluation metrics between different orders of data addition, with low differences represented by lighter colors and high differences by darker colors. The largest differences in precision and recall across the six data addition orders were 0.0193 and 0.0251, respectively. These variations are particularly noticeable in models such as U-Net, Res U-Net, and Dense U-Net. Notably, as the network model becomes more complex, the performance of the models trained with different data orders tends to stabilize. We think that when models have lower complexity and fewer parameters, there may have a mismatch between the capacity of the dataset and the model, leading to fluctuations in the training results.

Additionally, we observe that the F1-score difference across the performance of LMHLD$_{part}$ on seven models is minimal, with a maximum difference of only 0.0082. This indicates that the different sub-datasets all exhibit strong consistency and robustness, demonstrating that LMHLD exhibits stable performance when applied to various DL models for landslide detection. The average evaluation metrics, derived from training the models multiple times in different data addition orders, are summarized in Table 7. With the exception of Res U-Net, all models achieve high F1-scores, with the highest being 0.8699 for U-Net.

Comparing Table 7 with Table 6, we observe that the performance of LMHLD$_{part}$ on DL models for landslide detection is higher than that of LMHLD$_{mix}$. This outcome suggests that incorporating multiple patch sizes in the landslide dataset can improve landslide detection performance. Therefore, a qualified benchmark dataset for landslide detection should include optimal patch sizes. Additionally, comparing Table 7 with Table 5, we find that the average evaluation metrics of LMHLD$_{part}$ on the seven models surpass the performance of the network on other sub-datasets. This indicates that the DL model successfully learns new knowledge while minimizing the forgetting of prior knowledge. We believe LMHLD$_{part}$ effectively mitigates the issue of catastrophic forgetting. Furthermore, this illustrates that the LMHLD$_{part}$ module enhances the scalability of LMHLD, positioning it as a promising benchmark dataset for landslide detection.

Based on the aforementioned conclusions, LMHLD, when trained using the LMHLD$_{part}$ module, has the potential to incorporate more types of landslide data in future applications. This will allow DL models for landslide detection to leverage more comprehensive features and enhance their extraction capabilities.

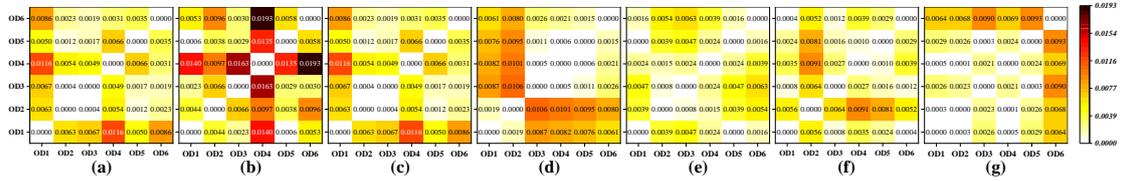

**Fig. 12** The precision of models trained with different orders of data addition. (a) U-Net. (b) Res U-Net. (c) Dense U-Net. (d) Attention U-Net. (e) Trans U-Net. (f) U-Net++. (g) U-Net+++. OD$n$ stands for the nth addition order ($n$ = 1, 2, 3, 4, 5, 6).

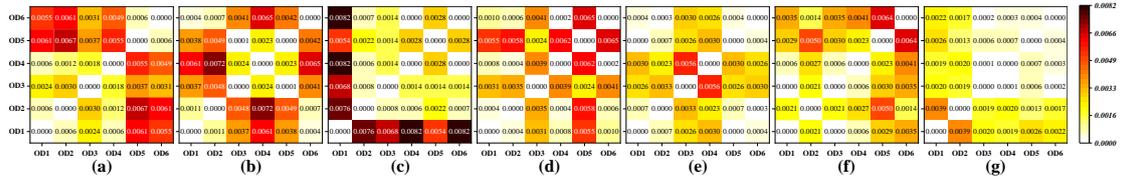

**Fig. 13** The f1-score of models trained with different orders of data addition. (a) U-Net. (b) Res U-Net. (c) Dense U-Net. (d) Attention U-Net. (e) Trans U-Net. (f) U-Net++. (g) U-Net+++. OD$n$ stands for the nth addition order ($n$ = 1, 2, 3, 4, 5, 6).

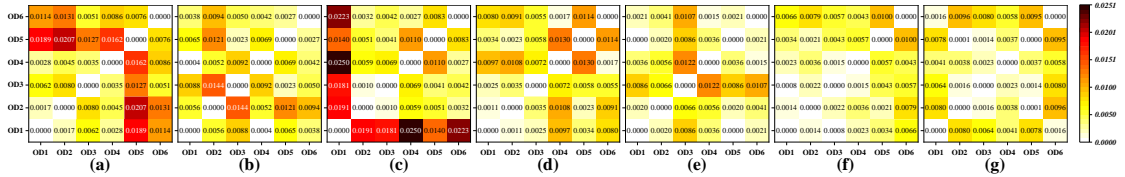

**Fig. 14** The recall of models trained with different orders of data addition. (a) U-Net. (b) Res U-Net. (c) Dense U-Net. (d) Attention U-Net. (e) Trans U-Net. (f) U-Net++. (g) U-Net+++. OD$n$ stands for the nth addition order ($n$ = 1, 2, 3, 4, 5, 6).

**Table 7** The performance of LMHLD$_{part}$ on seven deep learning models. The values are average evaluation metrics of the different orders of data addition.

| Model | Precision | F1-score | Recall |
| --- | --- | --- | --- |
| U-Net | **0.9083** | **0.8699** | 0.8346 |
| Res U-Net | 0.8704 | 0.8295 | 0.7926 |
| Dense U-Net | 0.8973 | 0.8623 | 0.8300 |
| Attention U-Net | 0.8612 | 0.8590 | 0.8568 |
| Trans U-Net | 0.9057 | 0.8646 | 0.8270 |
| U-Net++ | 0.8672 | 0.8663 | **0.8655** |
| U-Net+++ | 0.8776 | 0.8642 | 0.8513 |

It is important to note that the DL models from the U-Net family tested in this research are not state-of-the-art (SOTA). Specifically, the basic U-Net achieved the highest F1-score in the

LMHLD$_{part}$ experiment, as shown in Table 7. However, this does not necessarily indicate that U-Net delivers the best overall performance, but rather the highest evaluation index. As shown in Table 7, the difference between precision and recall for U-Net is 0.0737, which is larger than that for U-Net++ and U-Net+++. This suggests that U-Net is more conservative in predicting positive samples, potentially leading to missed detections of actual landslides in practical applications. For example, in Fig. 15, some pixels are missed in the landslide detection results, causing internal discontinuities in the detected landslides, a situation that should be avoided in real-world scenarios. In contrast, SOTA models, such as Trans U-Net and U-Net+++, demonstrate more stable performance.

Secondly, in the LMHLD$_{part}$ experiment, although U-Net achieved the highest F1 values, the performance gap between U-Net and other models, except for Res U-Net, is very small, with the maximum gap being only 0.0109. We think that in binary classification semantic segmentation tasks like landslide detection, if the dataset features are uniform, the performance differences among different models may be minimal. Basic models, such as U-Net, are already capable of capturing the essential data characteristics well, so more complex models like Trans U-Net or U-Net+++ may not yield significant performance improvements. Additionally, we used the same default values (e.g., hyperparameters, optimizer, learning rate) across all models, which may not fully exploit the potential of each model. Advanced models, in particular, may require more refined hyperparameter optimization to realize their full benefits

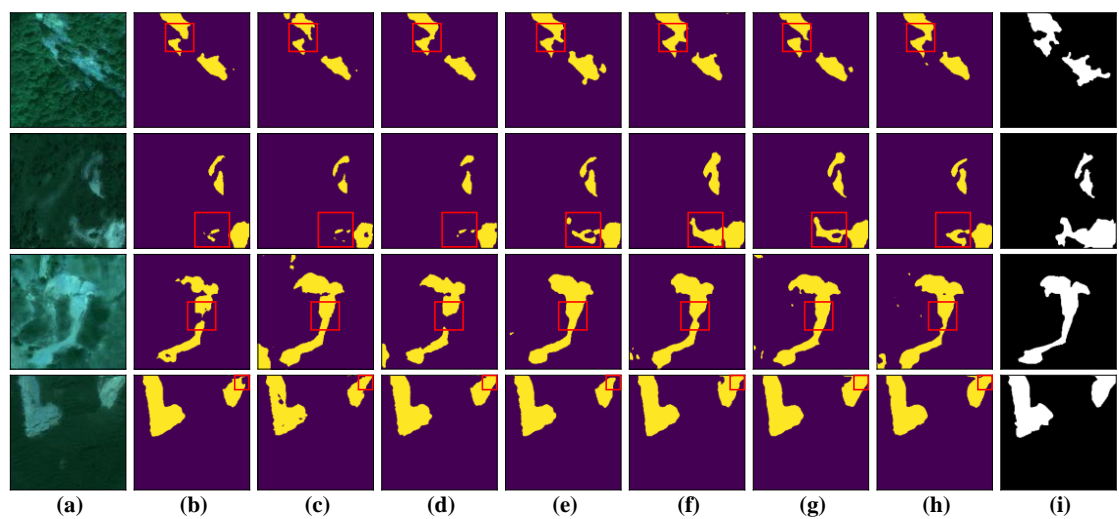

**Fig. 15 Landslide detection results of patches by the training module-LMHLD$_{part}$. (a) Original Image. (b) U-Net. (c) Res U-Net. (d) Dense U-Net. (e) Attention U-Net. (f) Trans U-Net. (g) U-Net++. (h) U-Net+++. (i) Ground Truth.**

### 4.5 Quality validation of LMHLD

We used U-Net, trained with the LMHLD$_{part}$, to detect landslides in images selected from the GVLM dataset and the recent landslide events in Luding, 2022. These images were completely unseen to the model. The landslide detection results are presented in Fig. 16 and Fig. 17, where red pixels represent the detected landslides. The results show that U-Net successfully identifies most landslides in the GVLM dataset, although a few roads are misclassified as landslides. Notably, the recent landslides caused by the earthquake in Luding were also detected by the model. Most large riverfront landslides are accurately identified and clearly differentiated from the river and its sediments. These findings demonstrate that the model trained with LMHLD$_{part}$ exhibits good generalization. Furthermore, LMHLD provides the DL models with sufficient landslide knowledge, enabling them to quickly produce reliable landslide inventories for new events.

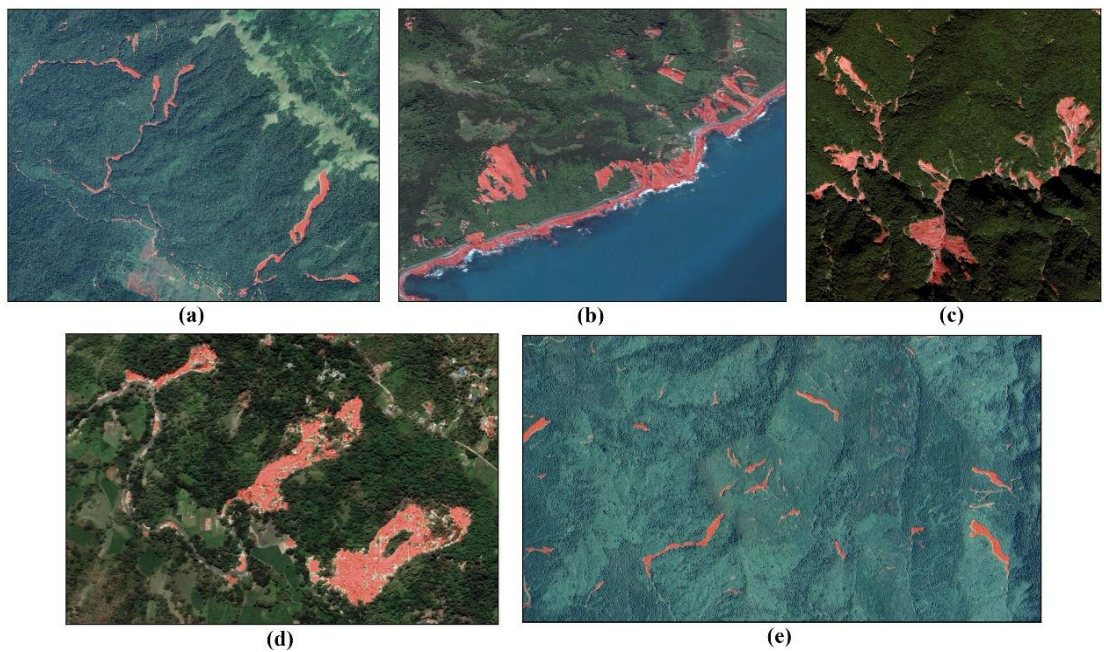

Fig. 16    Landslide detection results of the test data selected from GVLM. (a) Kodagu_India. (b) Kaikoura_New Zealand. (c) Taitung_China. (d) Kupang_Indonesia. (e) ALuoi_Vietnam.

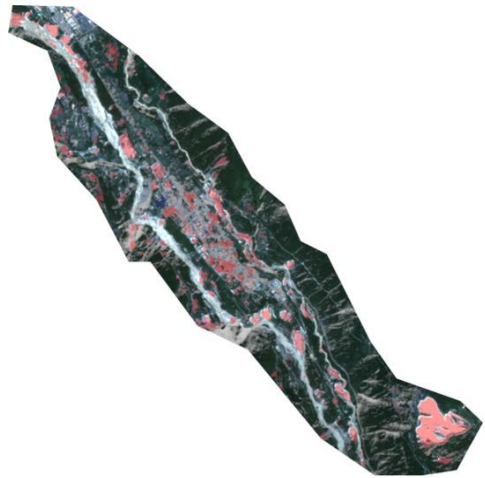

Fig. 17    Landslide detection results in Luding, China in 2022.

## 5 Conclusion

To effectively address sudden landslide disasters, it is crucial to quickly generate an accurate landslide inventory. Currently, many DL models are widely applied to landslide detection to achieve this objective. However, training data are essential for these models to learn landslide features, and the availability of high-quality, publicly labeled landslide datasets remains a significant limitation in the field. In this paper, we introduce LMHLD, a novel benchmark dataset specifically designed for landslide detection tasks using large-scale multi-source high-resolution remote sensing images. The proposed dataset boasts several notable features.

First, it leverages the strengths of recently published landslide datasets during its construction, covering a wide range of landslide events with diverse spatial distributions, triggering factors, terrain types, shapes, and sizes. Second, LMHLD integrates data from multiple satellite sensors and includes images of varying spatial resolutions, making it more adaptable to real-world scenarios. Finally, we have selected optimal patch sizes for landslides of different scales and images with varying spatial resolutions, ensuring that LMHLD provides rich and comprehensive features that can better support DL models in landslide detection tasks.

A reliable benchmark dataset for landslide detection must undergo rigorous quality testing before it can be released. To demonstrate that LMHLD has the potential to become a benchmark dataset for landslide detection, we designed five distinct experiments. These experiments highlight the dataset's robustness and its ability to support effective DL model training.

We also introduced a novel training module, $LMHLD_{part}$, which enables DL models to detect landslides across a broader range of scales. Additionally, $LMHLD_{part}$ helps mitigate the issue of catastrophic forgetting during multi-task learning, ensuring that the model can learn new knowledge without losing previously acquired knowledge. Furthermore, $LMHLD_{part}$ facilitates the seamless expansion and updating of multiple landslide datasets, significantly reducing the time required for the training process. This makes LMHLD not only a valuable resource for current landslide detection tasks but also a flexible and scalable tool for future research.

Although LMHLD is solely composed of remote sensing images, the DL models trained on it achieved outstanding landslide detection results on the GVLM dataset, demonstrating that LMHLD is both reliable and capable of providing excellent generalization for DL models. Moreover, the introduction and implementation of $LMHLD_{part}$ offer a significant advancement, enabling researchers to seamlessly integrate additional UAV data with LMHLD. This capability expands the dataset's versatility and enhances its potential for future landslide detection applications, further supporting the development of robust models capable of handling diverse data sources.

In summary, LMHLD is introduced as a benchmark dataset for landslide detection, offering comprehensive and high-quality training data for DL models. It also serves as a valuable evaluation criterion to assess the generalization ability of the latest DL models, advancing the progress of DL in this domain. Ultimately, we aim for more DL models trained with LMHLD to be deployed in real-world landslide disaster prevention and control, contributing significantly to the protection of lives and property.

## CRediT authorship contribution statement

**Guanting Liu:** Conceptualization, Data curation, Methodology, Validation, Formal analysis, Visualization, Writing – original draft, Writing – review & editing. **Yi Wang:** Conceptualization, Formal analysis, Writing – review & editing. **Xi Chen**: Data curation, Formal analysis, Writing – review & editing. **Zhice Fang**: Data curation, Writing – review & editing. **Baoyu Du**: Data curation, Writing – review & editing. **Penglei Li**: Data curation. **Yuan Wu**: Data curation.

## Declaration of Competing Interest

The authors declare that they have no known competing financial interests or personal relationships that could have appeared to influence the work reported in this paper.

## Acknowledgement

This work was supported by the Joint Funds of the National Natural Science Foundation of China (U21A2013), Key Project of the Open Fund of Key Laboratory of Ocean Space Resource

Management Technology (KF-2021-105).

## Data availability

The data and codes that support the findings of this study can be accessed at: https://doi.org/10.5281/zenodo.11424988.